\documentclass[conference]{IEEEtran}
\usepackage{cite}
\usepackage{amsmath,amssymb,amsfonts}
\usepackage{algorithmic}
\usepackage{graphicx}
\usepackage{textcomp}
\usepackage{multirow}
\usepackage{array}
\usepackage{booktabs}
\usepackage{bm}
\usepackage{tikz}
\usetikzlibrary{shapes.geometric, arrows, positioning, fit}

\usepackage{graphicx}
\usepackage{subcaption}

\usepackage{xcolor}

\definecolor{darkgreen}{rgb}{0.0, 0.5, 0.0}
\definecolor{darkred}{rgb}{0.55, 0.0, 0.0}
\definecolor{darkblue}{rgb}{0.0, 0.0, 0.55}
\definecolor{darkpurple}{rgb}{0.4, 0.0, 0.4}
\definecolor{darkbrown}{rgb}{0.4, 0.26, 0.13}
\definecolor{darkgoldenrod}{rgb}{0.72, 0.53, 0.04}
\definecolor{darkcyan}{rgb}{0.0, 0.4, 0.4}

\begin{document}
\title{DiSSECT: Structuring Transfer-Ready Medical Image Representations through Discrete Self-Supervision }

\author{Azad Singh, and Deepak Mishra, \IEEEmembership{Member, IEEE}
\thanks{ All the authors are with the Department of Computer Science, Indian Institute of Technology Jodhpur,
Jodhpur, Rajasthan 342204 India (e-mail: {\{singh.63,dmishra\}}@iitj.ac.in).}}

\maketitle

\begin{abstract}
Self-supervised learning (SSL) has emerged as a powerful paradigm for medical image representation learning, particularly in settings with limited labeled data. However, existing SSL methods often rely on complex architectures, anatomy-specific priors, or heavily tuned augmentations, which limit their scalability and generalizability. More critically, these models are prone to shortcut learning, especially in modalities like chest X-rays, where anatomical similarity is high and pathology is subtle. In this work, we introduce \textbf{DiSSECT}—\textbf{Di}screte \textbf{S}elf-\textbf{S}upervision for \textbf{E}fficient \textbf{C}linical \textbf{T}ransferable Representations, a framework that integrates multi-scale vector quantization into the SSL pipeline to impose a discrete representational bottleneck. This constrains the model to learn repeatable, structure-aware features while suppressing view-specific or low-utility patterns, improving representation transfer across tasks and domains. DiSSECT achieves strong performance on both classification and segmentation tasks, requiring minimal or no fine-tuning, and shows particularly high label efficiency in low-label regimes (1–5\%). We validate DiSSECT across multiple public medical imaging datasets, demonstrating its robustness and generalizability compared to existing state-of-the-art approaches.
\end{abstract}


\section{Introduction}
 Self-supervised learning (SSL) has emerged as a promising approach in medical imaging, enabling representation learning from unlabeled data through proxy tasks~\cite{jigsaw, colorization, rotpred, simclr, simsiam}. Among its techniques, contrastive learning and masked image modeling show strong potential, yet they often struggle to capture subtle abnormalities, adapt across domains, and generalize effectively, as they rely on superficial patterns rather than clinically meaningful cues~\cite{lin2024shortcut,hill2024risk}. This problem is particularly evident in chest X-rays, where global anatomical structures remain highly consistent across patients, while pathological variations such as lesions or nodules are sparse and localized. Pretext tasks emphasizing invariance or reconstruction therefore bias models toward dominant features like outlines or background gradients that persist across augmentations but lack diagnostic value. As a result, SSL representations frequently fail to support pathology-focused downstream tasks and require extensive fine-tuning, undermining the goal of label-efficient generalization and adding computational overhead in clinical deployment.

To improve semantic alignment, recent methods have introduced architectural complexity, anatomy-specific priors, or handcrafted augmentations~\cite{pcrlv1, transvw, dira, adamv2}. While effective, transformer-based models like DINO~\cite{dino} and MAE demand high-resolution inputs and long training, and anatomy-guided approaches (AnatPaste~\cite{sato2023anatomy}, SAM~\cite{sam}) rely on segmentation labels, atlas templates, or domain-specific logic. These strategies limit scalability and still require extensive fine-tuning, underscoring the need for a simpler, more generalizable SSL framework that avoids shortcut learning and yields stable, semantic representations.
We argue that this requires enforcing an information bottleneck that forces models to retain clinically relevant structure over superficial invariances. In chest X-rays, for example, global anatomy is consistent across patients, while pathology appears as subtle local variations; without a bottleneck, models default to uninformative cues like scan boundaries or positional patterns. Vector quantization (VQ) offers a natural solution by discretizing latent space into a finite codebook, compressing away view-specific noise while preserving robust features. Although VQ has shown promise in representation learning, its exploration in medical SSL is limited. Building on this insight and our prior work, CoBooM~\cite{coboom}, which introduced codebook-guided representation learning for medical images, we extend this direction toward a more general and scalable SSL framework.

In this work, we present DiSSECT (\textbf{Di}screte \textbf{S}elf-\textbf{S}upervision for \textbf{E}fficient \textbf{C}linical \textbf{T}ransferable Representations), a simple yet general framework for learning transferable medical image representations through multi-scale discrete supervision. Unlike prior methods that depend on anatomy-specific labels, multi-branch objectives, or handcrafted augmentations, DiSSECT integrates lightweight discrete bottleneck modules at multiple semantic levels during pretraining. This enforces structured and repeatable features that generalize across tasks, modalities, and datasets, yielding an organized latent space that supports strong classification and segmentation performance under minimal or no fine-tuning. We validate DiSSECT on NIH ChestX-ray14, CheXpert, RSNA Pneumonia, and SIIM-ACR Pneumothorax, demonstrating robust performance in low-label regimes.
Our key contributions are:
\begin{itemize}
\item Introducing DiSSECT, a discrete SSL framework that leverages multi-scale VQ to enforce an information bottleneck.
\item Showing that VQ-enhanced SSL improves representation structure and stability, enabling effective transfer with little or no fine-tuning—critical in low-resource clinical settings.
\item Providing comprehensive evaluations across multiple benchmarks, where DiSSECT achieves competitive or superior performance in classification and segmentation under limited labels.
\end{itemize}

\section{Related Works}
\subsection{Self-Supervised Learning in Medical Imaging }
Recent advances in SSL have shown promise in medical imaging, particularly contrastive learning~\cite{simclr,simclrv2,mocov2}. Extensions include adapting to 3D data~\cite{chaitanya2020contrastive}, leveraging patient metadata~\cite{dufumier2021contrastive}, and scaling to large datasets~\cite{ghesu2022contrastive}. Hybrid approaches such as PCRLv1/v2~\cite{pcrlv1,pcrlv2} and DiRA~\cite{dira} combined contrastive and restorative (or adversarial) losses, improving diversity but at the cost of multi-stage training and high computation. Anatomy-guided methods (e.g., TransVW~\cite{transvw}, ADAM~\cite{adamv1,adamv2}, SAM~\cite{sam}) incorporated structural priors via alignment, hierarchies, or atlas templates, boosting semantic consistency but limiting scalability due to template dependence and dense comparisons.
Transformer-based SSL has also emerged: DISTL~\cite{distl} iteratively refined representations via self-distillation, DASQE~\cite{DASQE} disentangled anatomy from artifacts for domain adaptation, and UniMiSS+~\cite{unimiss} learned modality-agnostic features across 2D/3D data. Chest X-ray–focused methods include MLVICX~\cite{mlvicx}, emphasizing multi-level invariance, OTCXR~\cite{otcxr}, leveraging optimal transport for semantic alignment, and AFiRe~\cite{afire}, aligning ViT tokens with radiographic structures. Despite these advances, most methods remain constrained by domain-specific assumptions, multi-stage pipelines, or high computational cost, underscoring the need for a generalizable and efficient SSL framework.

\begin{figure*}[t]
\centering
\includegraphics[width=1.0\textwidth]{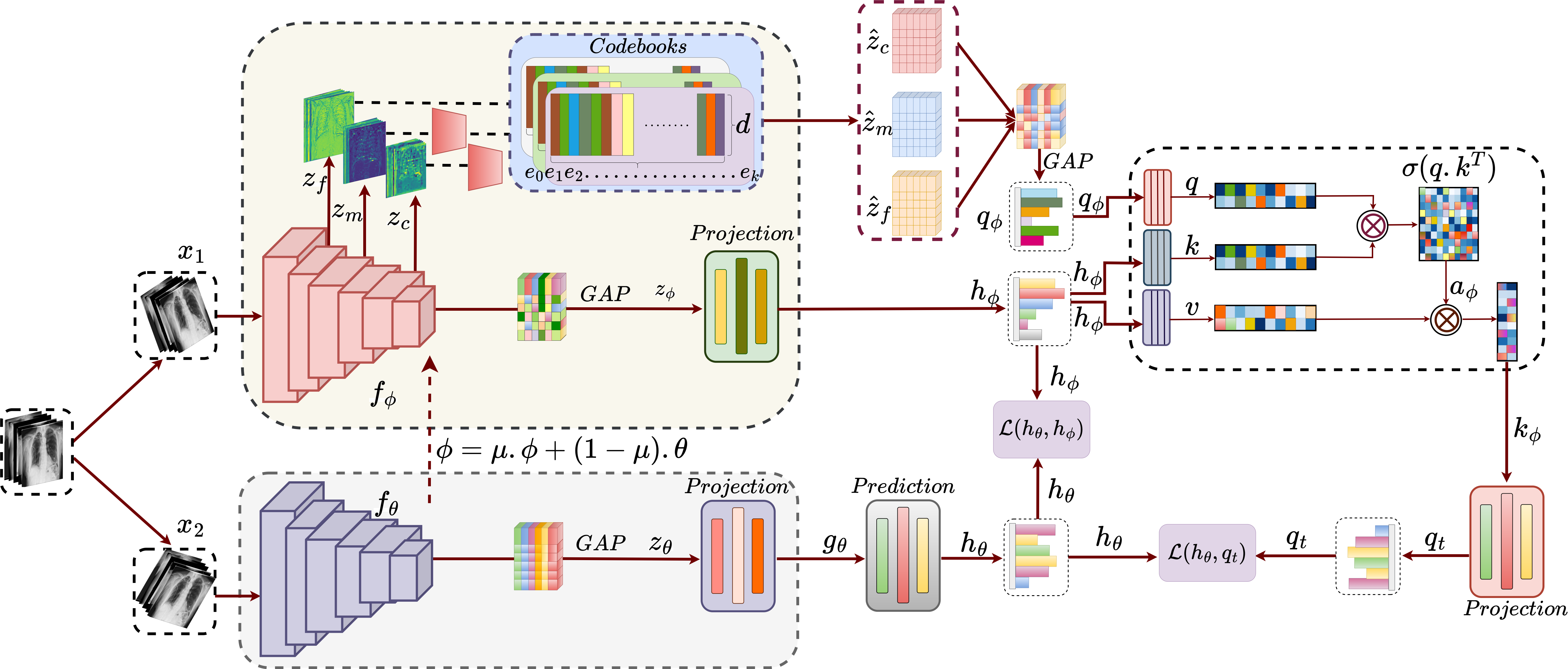}
\caption{DiSSECT architecture. Two augmented views are processed by a representation branch ($f_\theta$) and a momentum-updated supervision branch ($f_\phi$). Multi-scale features ($z_c$, $z_m$, $z_f$) from $f_\phi$ are quantized via dedicated codebooks and fused to form $q_\phi$. Cross-attention with the global feature $h_\phi$ yields $k_\phi$, projected to produce the supervision target $q_t$. The encoder $f_\theta$ is trained using dual alignment losses: to $h_\phi$ and $q_t$, promoting semantic and structural consistency.
}
\label{main_fig}
\end{figure*}

\subsection{Vector Quantization and Discrete Representation Learning in Medical Imaging}
Vector quantization (VQ) has re-emerged as a powerful tool for enforcing discrete inductive biases in representation learning. Originally popularized in generative models like VQ-VAE~\cite{vqvae} and VQ-GAN~\cite{vqgan}, VQ maps continuous features to codebook entries, yielding structured and compressible representations. Beyond speech modeling~\cite{wave2vec,hubert}, approaches such as BEiT~\cite{bao2022beit} and MaskGIT~\cite{chang2022maskgit} have demonstrated their utility in vision via patch-level quantization, motivating exploration in medical imaging.
In this domain, VQ applications remain nascent. Prior work has applied VQ to improve segmentation robustness~\cite{santhirasekaram2023topology,santhirasekaram2023sheaf,wang2025lightweight,huang2023rethinking}, generative modeling~\cite{zhou2025generating,han2024non}, and self-supervised instance discrimination~\cite{idqce}. However, integrating VQ into SSL is non-trivial, facing challenges such as codebook collapse, gradient instability, and underutilization without strong supervision. Our prior work, CoBooM~\cite{coboom}, combined discrete and continuous cues, showing promise for anatomy-aware learning. Building on this, DiSSECT introduces multi-scale VQ across semantic levels within a unified encoder, enforcing structure without handcrafted augmentations or anatomical labels. Unlike VQ-VAE–based generative models, DiSSECT applies quantization purely for discriminative representation learning, producing efficient, transferable, and anatomically structured embeddings with minimal downstream supervision.

\section{Method}
DiSSECT, illustrated in Figure~\ref{main_fig}, consists of two branches: a representation branch that learns the primary encoder, and a discrete supervision branch that integrates multi-scale VQ modules to provide quantized, structure-aware targets. The supervision branch discretizes intermediate feature maps at coarse, medium, and fine resolutions using dedicated codebooks, fuses them, and produces stable targets. This encourages the encoder to capture compressible, clinically meaningful patterns while suppressing view-specific artifacts and overly dominant anatomical structures. Without using pathology labels, the discrete bottleneck biases representations toward localized, repeatable deviations from typical anatomy—often aligned with subtle abnormalities.

\textbf{Representation Branch and Feature Embedding:} The representation branch ($f_\theta$) learns transferable features from augmented views ($x_1$, $x_2$). Each view is passed through a shared convolutional encoder to yield feature maps $z_\theta$, which are pooled and projected via an MLP into latent vectors $g_\theta$. A lightweight prediction head then produces $h_\theta$, trained to align with supervisory signals from the discrete supervision branch ($f_\phi$). Supervision is applied through two complementary signals: one from a continuous global embedding, ensuring high-level semantic coherence, and another from a multi-scale, structurally refined, quantized pathway that emphasizes compressible anatomical patterns and localized pathological cues. This dual-target strategy mitigates shortcut learning by constraining representations to be both semantically rich and structurally refined. As a result, the encoder learns stable and generalizable features applicable across medical imaging tasks.

\textbf{Discrete Supervision Branch: } 
The discrete supervision branch ($f_\phi$) complements the representation branch by generating structure-aware supervisory signals. It processes the same augmented views as the representation branch, using a shared encoder to extract features at three semantic levels, coarse ($z_c$), medium ($z_m$), and fine ($z_f$), capturing both global context and localized anatomical details. A global average-pooled embedding from the final layer is projected to produce a continuous target vector $h_\phi$. Unlike $f_\theta$, the parameters of $f_\phi$ are updated via momentum averaging ($\phi \leftarrow \mu \cdot \phi + (1 - \mu) \cdot \theta$), ensuring stable targets during training, where $\mu$ is the momentum coefficient. Together, the continuous target $h_\phi$ and quantized multi-scale features ($z_c$, $z_m$, $z_f$) form the basis of DiSSECT’s dual supervision strategy. The specific construction of the quantized supervisory pathway is described in the following section.

\textbf{Rationale for Quantization Operates on the Momentum-Updated Branch:}
In DiSSECT, quantization operates on the momentum-updated encoder $f_\phi$ rather than the gradient-updated encoder $f_\theta$. The rationale is rooted in vector quantization theory: stable encoder outputs are critical for effective codebook evolution, as VQ updates rely on exponential moving averages (EMA) of assigned features. Formally, the EMA update rule for codeword $\mathbf{e}_i$ at iteration $t$ is:
\begin{equation}
\mathbf{e}_i^{(t+1)} = m \cdot \mathbf{e}_i^{(t)} + (1 - m) \cdot \bar{z}_i^{(t)},
\end{equation}
where $m \in [0,1)$ is the momentum coefficient, and $\bar{z}_i^{(t)}$ is the mean of encoder features $\tilde{z}$ assigned to codeword $i$ at step $t$. The incremental update is:
\begin{equation}
\Delta \mathbf{e}_i^{(t)} = (1 - m)(\bar{z}_i^{(t)} - \mathbf{e}_i^{(t)}).
\end{equation}
Thus, the variance of the update is:
\begin{equation}
\mathrm{Var}[\Delta \mathbf{e}_i^{(t)}] = (1 - m)^2 \cdot \mathrm{Var}[\bar{z}_i^{(t)}].
\end{equation}
This reveals a key insight: if encoder outputs $\tilde{z}$ shift rapidly (as in $f_\theta$ during gradient descent), then $\mathrm{Var}[\bar{z}i^{(t)}]$ is high, leading to noisy, unstable codebook updates. Such instability can cause poor utilization of codewords (collapse) and degrade the quality of discrete representations. In contrast, $f\phi$, updated slowly via momentum averaging, yields smoother features across iterations, reducing $\mathrm{Var}[\Delta \mathbf{e}i^{(t)}$], promoting stable codebook convergence, and ensuring consistent use of discrete entries. Empirically (Section VI), quantizing $f\theta$ leads to training collapse, while $f_\phi$ enables robust convergence.
Importantly, stable codebooks allow DiSSECT to capture repeatable and compressible anatomical structures, rather than drifting toward view-specific noise or unstable patterns. This stability is essential for reliable medical representations, where consistency across patients and scans is critical.

\textbf{Multi-Scale Feature Quantization:}
\label{sec:multi-scale-vq} DiSSECT applies quantization separately on feature maps $z_j \in \mathbb{R}^{C_j \times H_j \times W_j}$, extracted from the supervision branch, at the scale $j \in {c, m, f}$, where $C_j$ is the number of channels and $H_j \times W_j$ is the spatial resolution. Each scale is associated with its respective learnable codebook 
$\mathcal{E}_j \in \mathbb{R}^{N_j \times d}$. $N_j$ denotes the number of codewords in the codebook for $j^{th}$ scale feature maps, each denoted as $e^n_j \in \mathbb{R}^d$ where $d$ is dimension of each codeword and $n$ denotes the $n^{th}$ codebook vector in $\mathcal{E}_j$. Before quantization, each feature map $z_j$ is projected to a common embedding space of dimension $d$ using a $1{\times}1$ convolution and reshaped into a sequence of spatial patch-level vectors $\tilde{z}_j \in \mathbb{R}^{(H_j \cdot W_j) \times d}$. Each vector $\tilde{z}_j[p]$ is then assigned to its nearest codeword $e_j$ in $\mathcal{E}_j$ via:

\begin{equation}
\hat{z}_j[p] = \mathbf{e}_j^{n^*}; \quad n^* = \operatorname*{arg\,min}_{n} \left\| \tilde{z}_j[p] - \mathbf{e}_j^n \right\|_2^2
\label{eq:vq}
\end{equation}
Here, $n^*$ denotes the index of the optimal codeword in the codebook. 

This yields a quantized map $\hat{z}_j \in \mathbb{R}^{d \times H_j \times W_j}$ for each scale, preserving spatial structure. The VQ process acts as a discrete bottleneck that suppresses non-discriminative patterns and dominant anatomical redundancies, while emphasizing structurally compressible, repeatable features. This biases the model toward localized deviations that often align with clinically relevant abnormalities, even without explicit labels.
The quantized outputs $\hat{z}_c$, $\hat{z}_m$, and $\hat{z}_f$ are later used to construct the multi-scale discrete supervisory signal.
To stabilize training and encourage codebook usage, we apply a commitment loss for each scale:
\begin{equation}
L_{\text{VQ}, j} = \beta \cdot \left| \tilde{z}_j - \text{sg}[\hat{z}_j] \right|_2^2,
\label{eq:vq_loss}
\end{equation}
where $\text{sg}[\cdot]$ is the stop-gradient operator and $\beta$ controls the penalty strength. The total quantization loss is: $\mathcal{L}_{\text{VQ}} = \sum_{j \in {c, m, f}} L_{\text{VQ}, j}$. This multi-scale quantization encourages consistent, spatially grounded representations critical for robust medical feature learning.

\textbf{Structured Embedding Refinement Fusion (SERF)}
\label{sec:serf}
Continuous embeddings ($h_\phi$) capture global semantic information, while quantized features ($\hat{z}_c, \hat{z}_m, \hat{z}_f$) emphasize localized, structurally repeatable patterns. In DiSSECT, we regard these as complementary views and fuse them to construct a refined supervision signal that balances semantic richness with structural stability. Specifically, quantized maps from coarse, medium, and fine scales ($\hat{z}_c$, $\hat{z}_m$, and $\hat{z}_f$ respectively) are projected into a shared space and combined via weighted fusion: $q_\phi = \alpha_c \cdot \hat{z}_c + \alpha_m \cdot \hat{z}_m + \alpha_f \cdot \hat{z}_f$, where $\alpha_c, \alpha_m, \alpha_f$ are scalar weights. After fusion of quantized maps, the pooled vector $q_\phi$ acts as a quantized anchor representing stable and compressible anatomical structure. This anchor is used to refine the global embedding $h_\phi \in \mathbb{R}^d$ through a cross-attention–inspired operator: 
\begin{equation}
k_\phi = \text{softmax}!\left( \frac{q_\phi h_\phi^\top}{\sqrt{d}} \right) h_\phi,
\end{equation}
Here, $q_\phi$ serves as the query, while $h_\phi$ acts as both key and value. By structuring the interaction in this way, refinement is explicitly grounded in discrete codes. This design provides several advantages: (i) Stability: $q_\phi$ comes from quantized features, which evolve slowly and consistently across training, anchoring the supervision signal. (ii) Semantic alignment: $h_\phi$ retains global semantic context; aligning it with $q_\phi$ prevents collapse into purely anatomical regularities. (iii) Clinical relevance: The fused vector $k_\phi$ reflects localized, compressible structures emphasized by quantization, while also retaining semantic variation that may correspond to pathological cues. If the refinement operator were omitted (e.g., relying only on $h_\phi$), supervision would risk overfitting to global but clinically uninformative signals. Conversely, using only quantized codes would lose broader semantic context, leading to fragmented features. SERF’s refinement balances these extremes by blending discrete anchors with continuous context. Finally, $k_\phi$ is passed through an MLP to obtain the SERF supervision target: $q_t = \text{MLP}(k_\phi) \in \mathbb{R}^d$.

% By anchoring attention on discretized features extracted from coarse, medium, and fine levels, CodeX produces a supervision signal that helps the model generalize across imaging variations and reduces dependence on extensive fine-tuning. The quantized feature maps $\hat{z}_c$, $\hat{z}_m$, and $\hat{z}_f$ are first projected into a shared space and spatially pooled if needed and are then fused via a weighted sum: $q_\phi = \alpha_c \cdot \hat{z}_c + \alpha_m \cdot \hat{z}_m + \alpha_f \cdot \hat{z}_f$ where $\alpha_c$, $\alpha_m$, and $\alpha_f$ are fixed scalar weights. 

% Applying global average pooling yields 

% The quantized anchor $q_\phi \in \mathbb{R}^d$, used as the query in a cross-attention mechanism with the continuous embedding $h_\phi \in \mathbb{R}^d$ as both key and value. This yields an attention-refined vector $k_\phi \in \mathbb{R}^d$,: 
% \begin{equation}
% k_\phi = \text{softmax}\left( \frac{q_\phi h_\phi^\top}{\sqrt{d}} \right) h_\phi.
% \end{equation}
% Using $q_\phi$ as the query grounds the attention in stable, discrete patterns that are anatomically compressible and consistent, guiding the model to focus on semantically meaningful regions in $h_\phi$. The result, $k_\phi$, blends structural and contextual cues, enhancing robustness and transferability. 

% Finally, $k_\phi$ is passed through an MLP projection head to obtain the discrete-refined supervision target: $q_t = \text{MLP}(k_\phi) \in \mathbb{R}^d.$

\begin{table*}[htbp]
\centering
\caption{SSL performance on NIH (original split) under FT and LP across label proportions (1–40\%). Each column shows FT, LP, and LP–FT difference ($\Delta$).
Best and second-best scores are marked in \textcolor{darkblue}{\textbf{blue}} and \textcolor{darkpurple}{\textbf{purple}}; positive and negative $\Delta$ in \textcolor{darkgreen}{\textbf{green}} and \textcolor{darkred}{\textbf{red}}.}

\label{tab:nih_orig}
\setlength{\tabcolsep}{4pt}
\begin{tabular}{l|ccc|ccc|ccc|ccc|ccc|ccc}
\toprule
\multirow{2}{*}{\textbf{Method}} 
& \multicolumn{3}{c|}{\textbf{1\%}} 
& \multicolumn{3}{c|}{\textbf{5\%}} 
& \multicolumn{3}{c|}{\textbf{10\%}} 
& \multicolumn{3}{c|}{\textbf{20\%}} 
& \multicolumn{3}{c|}{\textbf{30\%}} 
& \multicolumn{3}{c}{\textbf{40\%}} \\
 & FT & LP & $\Delta$ & FT & LP & $\Delta$ & FT & LP & $\Delta$ & FT & LP & $\Delta$ & FT & LP & $\Delta$ & FT & LP & $\Delta$ \\
\midrule
Random Init.     & 57.7 & 54.7 & \textbf{\textcolor{darkred}{-3.0}} & 62.7 & 57.4 & \textbf{\textcolor{darkred}{-5.3}} & 65.6 & 59.0 & \textbf{\textcolor{darkred}{-6.6}} & 67.2 & 61.0 & \textbf{\textcolor{darkred}{-6.2}} & 70.7 & 61.1 & \textbf{\textcolor{darkred}{-9.6}} & 71.0 & 61.4 & \textbf{\textcolor{darkred}{-9.6}} \\
ImageNet Init.   & 61.0 & 59.1 & \textbf{\textcolor{darkred}{-1.9}} & 65.7 & 64.0 & \textbf{\textcolor{darkred}{-1.7}} & 68.2 & 65.1 & \textbf{\textcolor{darkred}{-3.1}} & 71.5 & 67.0 & \textbf{\textcolor{darkred}{-4.5}} & 73.9 & 67.6 & \textbf{\textcolor{darkred}{-6.3}} & 75.3 & 68.0 & \textbf{\textcolor{darkred}{-7.3}} \\
TransVW~\cite{transvw} & 62.3 & 59.4 & \textbf{\textcolor{darkred}{-2.9}} & 66.1 & 63.9 & \textbf{\textcolor{darkred}{-2.2}} & 68.8 & 64.3 & \textbf{\textcolor{darkred}{-4.5}} & 71.9 & 67.8 & \textbf{\textcolor{darkred}{-4.1}} & 74.3 & 68.1 & \textbf{\textcolor{darkred}{-6.2}} & 75.5 & 68.9 & \textbf{\textcolor{darkred}{-6.6}} \\
BYOL~\cite{byol}         & 64.2 & 63.1 & \textbf{\textcolor{darkred}{-1.1}} & 68.4 & 67.0 & \textbf{\textcolor{darkred}{-1.4}} & 70.7 & 69.3 & \textbf{\textcolor{darkred}{-1.4}} & 73.3 & 71.4 & \textbf{\textcolor{darkred}{-1.9}} & 75.3 & 71.9 & \textbf{\textcolor{darkred}{-3.4}} & 75.7 & 72.8 & \textbf{\textcolor{darkred}{-2.9}} \\
SimSiam~\cite{simsiam}   & 61.4 & 61.0 & \textbf{\textcolor{darkred}{-0.4}} & 67.0 & 64.3 & \textbf{\textcolor{darkred}{-2.7}} & 69.5 & 66.4 & \textbf{\textcolor{darkred}{-3.1}} & 71.5 & 68.2 & \textbf{\textcolor{darkred}{-3.3}} & 72.9 & 68.7 & \textbf{\textcolor{darkred}{-4.2}} & 73.4 & 69.1 & \textbf{\textcolor{darkred}{-4.3}} \\
DIRA~\cite{dira}         & 63.1 & 60.1 & \textbf{\textcolor{darkred}{-3.0}} & 68.5 & 65.5 & \textbf{\textcolor{darkred}{-3.0}} & 71.1 & 67.8 & \textbf{\textcolor{darkred}{-3.3}} & 74.4 & 71.4 & \textbf{\textcolor{darkred}{-3.0}} & 76.7 & 72.1 & \textbf{\textcolor{darkred}{-4.6}} & 77.4 & 72.8 & \textbf{\textcolor{darkred}{-4.6}} \\
PCRLv2~\cite{pcrlv2}     & 63.3 & 62.2 & \textbf{\textcolor{darkred}{-1.1}} & 69.8 & 66.6 & \textbf{\textcolor{darkred}{-3.2}} & 72.0 & 68.5 & \textbf{\textcolor{darkred}{-3.5}} & 74.7 & 71.2 & \textbf{\textcolor{darkred}{-3.5}} & 76.6 & 71.8 & \textbf{\textcolor{darkred}{-4.8}} & 76.9 & 72.0 & \textbf{\textcolor{darkred}{-4.9}} \\
DINOV2~\cite{dino}       & 61.3 & 59.6 & \textbf{\textcolor{darkred}{-1.7}} & 65.9 & 63.8 & \textbf{\textcolor{darkred}{-2.1}} & 68.1 & 64.6 & \textbf{\textcolor{darkred}{-3.5}} & 72.1 & 68.2 & \textbf{\textcolor{darkred}{-3.9}} & 74.6 & 68.7 & \textbf{\textcolor{darkred}{-5.9}} & 75.9 & 69.2 & \textbf{\textcolor{darkred}{-6.7}} \\
MLVICX~\cite{mlvicx}     & 65.6 & 66.3 & \textbf{\textcolor{darkgreen}{+0.7}} & 71.8 & 69.8 & \textbf{\textcolor{darkred}{-2.0}} & 73.3 & 70.6 & \textbf{\textcolor{darkred}{-2.7}} & 75.5 & 71.9 & \textbf{\textcolor{darkred}{-3.6}} & 76.2 & 73.0 & \textbf{\textcolor{darkred}{-3.2}} & 77.3 & 73.6 & \textbf{\textcolor{darkred}{-3.7}} \\
AdamV2~\cite{adamv2}     & 63.0 & 60.3 & \textbf{\textcolor{darkred}{-2.7}} & 68.1 & 65.5 & \textbf{\textcolor{darkred}{-2.6}} & 71.2 & 68.8 & \textbf{\textcolor{darkred}{-2.4}} & 74.2 & 71.4 & \textbf{\textcolor{darkred}{-2.8}} & 76.0 & 72.3 & \textbf{\textcolor{darkred}{-3.7}} & 76.4 & 72.8 & \textbf{\textcolor{darkred}{-3.6}} \\
OTCXR~\cite{otcxr}       & 65.9 & 61.2 & \textbf{\textcolor{darkred}{-4.7}} & 72.2 & 64.3 & \textbf{\textcolor{darkred}{-7.9}} & 73.9 & 67.6 & \textbf{\textcolor{darkred}{-6.3}} & 75.3 & 70.5 & \textbf{\textcolor{darkred}{-4.8}} & 76.1 & 71.8 & \textbf{\textcolor{darkred}{-4.3}} & \textbf{\textcolor{darkpurple}{77.4}} & 72.7 & \textbf{\textcolor{darkred}{-4.7}} \\
CoBoom~\cite{coboom}     & \textbf{\textcolor{darkpurple}{66.1}} & \textbf{\textcolor{darkpurple}{67.2}} & \textbf{\textcolor{darkgreen}{+1.1}} & \textbf{\textcolor{darkpurple}{72.4}} & \textbf{\textcolor{darkpurple}{71.7}} & \textbf{\textcolor{darkred}{-0.7}} & \textbf{\textcolor{darkpurple}{74.2}} & \textbf{\textcolor{darkpurple}{73.0}} & \textbf{\textcolor{darkred}{-1.2}} & \textbf{\textcolor{darkblue}{76.0}} & \textbf{\textcolor{darkpurple}{74.3}} & \textbf{\textcolor{darkred}{-1.7}} & \textbf{\textcolor{darkblue}{77.2}} & \textbf{\textcolor{darkpurple}{74.8}} & \textbf{\textcolor{darkred}{-2.4}} & \textbf{\textcolor{darkblue}{77.8}} & \textbf{\textcolor{darkpurple}{75.0}} & \textbf{\textcolor{darkred}{-2.8}} \\
\textbf{DiSSECT (Ours)}  & \textbf{\textcolor{darkblue}{66.8}} & \textbf{\textcolor{darkblue}{68.8}} & \textbf{\textcolor{darkgreen}{+2.0}} & \textbf{\textcolor{darkblue}{72.6}} & \textbf{\textcolor{darkblue}{72.7}} & \textbf{\textcolor{darkgreen}{+0.1}} & \textbf{\textcolor{darkblue}{74.8}} & \textbf{\textcolor{darkblue}{73.8}} & \textbf{\textcolor{darkred}{-1.0}} & \textbf{\textcolor{darkpurple}{75.8}} & \textbf{\textcolor{darkblue}{74.8}} & \textbf{\textcolor{darkred}{-1.0}} & \textbf{\textcolor{darkpurple}{76.8}} & \textbf{\textcolor{darkblue}{75.3}} & \textbf{\textcolor{darkred}{-1.5}} & 77.3 & \textbf{\textcolor{darkblue}{75.8}} & \textbf{\textcolor{darkred}{-1.5}} \\
\bottomrule
\end{tabular}
\end{table*}

\textbf{Training Strategy:} DiSSECT trains the encoder by aligning the prediction vector $h_\theta$ with two complementary targets: the global continuous embedding $h_\phi$ and the SERF-refined supervision signal $q_t$. This dual-target alignment ensures that representations capture both semantic context and discrete structural cues. To enforce alignment, we adopt a cosine regression loss:
\begin{equation}
\ell_{\text{reg}}(x, y) = 2 - 2 \cdot \frac{\langle x, y \rangle}{\|x\|_2 \cdot \|y\|_2},
\end{equation}
where $x, y \in \mathbb{R}^d$ are L2-normalized vectors. Cosine regression penalizes angular deviations, encouraging directional similarity in the embedding space and making the learned representations invariant to magnitude scaling.The alignment objective averages the two regression terms: $\mathcal{L}_{\text{sim}} = \frac{1}{2} \left( \ell_{\text{reg}}(h_\theta, h_\phi) + \ell_{\text{reg}}(h_\theta, q_t) \right)$.
Finally, the overall training objective combines this alignment loss with the multi-scale quantization commitment loss: $\mathcal{L}_{\text{total}} = \mathcal{L}_{\text{sim}} + \lambda \cdot \mathcal{L}_{\text{VQ}}$, 
where $\lambda$ controls the trade-off between semantic alignment and quantization stability. This joint supervision enforces that the encoder learns semantically meaningful yet structurally grounded representations, mitigating shortcut learning and reducing dependence on extensive fine-tuning. In practice, this design enables DiSSECT to generalize robustly across tasks and datasets, especially under low-label conditions.

%%%%%%%%%%%%%%%%%%%%%%%%%%%%%%%%%%%%%%%%%%%%%%%%%%%%%%%%%%%%%%%%%%%%%%%%%%%%%%%%%%%

\begin{table*}[htbp]
\centering
\caption{SSL performance on NIH (baseline split) under FT and LP across label fractions. Best and second-best FT/LP scores are marked in \textcolor{darkblue}{\textbf{blue}} and \textcolor{darkpurple}{\textbf{purple}}; LP–FT gaps in \textcolor{darkgreen}{\textbf{green}} (positive) and \textcolor{darkred}{\textbf{red}} (negative).}
\label{tab:nih_pcrl}
\setlength{\tabcolsep}{4pt}
\begin{tabular}{l|ccc|ccc|ccc|ccc|ccc|ccc}
\toprule
\multirow{2}{*}{\textbf{Method}} 
& \multicolumn{3}{c|}{\textbf{1\%}} 
& \multicolumn{3}{c|}{\textbf{5\%}} 
& \multicolumn{3}{c|}{\textbf{10\%}} 
& \multicolumn{3}{c|}{\textbf{20\%}} 
& \multicolumn{3}{c|}{\textbf{30\%}} 
& \multicolumn{3}{c}{\textbf{40\%}} \\
 & FT & LP & $\Delta$ & FT & LP & $\Delta$ & FT & LP & $\Delta$ & FT & LP & $\Delta$ & FT & LP & $\Delta$ & FT & LP & $\Delta$ \\
\midrule
Random Init.     & 58.9 & 56.6 & \textbf{\textcolor{darkred}{-2.3}} & 65.8 & 60.3 & \textbf{\textcolor{darkred}{-5.5}} & 68.1 & 63.1 & \textbf{\textcolor{darkred}{-5.0}} & 71.5 & 64.2 & \textbf{\textcolor{darkred}{-7.3}} & 73.4 & 64.6 & \textbf{\textcolor{darkred}{-8.8}} & 75.4 & 65.0 & \textbf{\textcolor{darkred}{-10.4}} \\
ImageNet Init.   & 63.9 & 62.2 & \textbf{\textcolor{darkred}{-5.0}} & 70.4 & 65.4 & \textbf{\textcolor{darkred}{-5.3}} & 73.5 & 68.2 & \textbf{\textcolor{darkred}{-5.6}} & 76.2 & 70.6 & \textbf{\textcolor{darkred}{-7.4}} & 78.5 & 71.1 & \textbf{\textcolor{darkred}{-5.0}} & 79.0 & 71.9 & \textbf{\textcolor{darkred}{-5.4}} \\
TransVW~\cite{transvw}   & 63.4 & 61.6 & \textbf{\textcolor{darkred}{-1.8}} & 66.5 & 64.9 & \textbf{\textcolor{darkred}{-1.6}} & 70.2 & 67.7 & \textbf{\textcolor{darkred}{-2.5}} & 74.3 & 70.3 & \textbf{\textcolor{darkred}{-4.0}} & 76.7 & 72.4 & \textbf{\textcolor{darkred}{-4.3}} & 77.6 & 73.7 & \textbf{\textcolor{darkred}{-3.9}} \\
BYOL~\cite{byol}         & 68.1 & 65.7 & \textbf{\textcolor{darkred}{-2.4}} & 73.2 & 71.5 & \textbf{\textcolor{darkred}{-1.7}} & 76.0 & 73.6 & \textbf{\textcolor{darkred}{-2.4}} & 76.8 & 74.8 & \textbf{\textcolor{darkred}{-2.0}} & 78.8 & 76.7 & \textbf{\textcolor{darkred}{-2.1}} & 79.8 & 77.3 & \textbf{\textcolor{darkred}{-2.5}} \\
SimSiam~\cite{simsiam}   & 66.2 & 64.6 & \textbf{\textcolor{darkred}{-1.6}} & 71.7 & 70.0 & \textbf{\textcolor{darkred}{-1.7}} & 74.0 & 72.0 & \textbf{\textcolor{darkred}{-2.0}} & 76.5 & 74.0 & \textbf{\textcolor{darkred}{-2.5}} & 78.0 & 75.0 & \textbf{\textcolor{darkred}{-3.0}} & 79.4 & 76.2 & \textbf{\textcolor{darkred}{-3.2}} \\
DIRA~\cite{dira}         & 66.6 & 62.3 & \textbf{\textcolor{darkred}{-4.3}} & 72.1 & 70.4 & \textbf{\textcolor{darkred}{-1.7}} & 74.6 & 73.8 & \textbf{\textcolor{darkred}{-0.8}} & 76.3 & 74.8 & \textbf{\textcolor{darkred}{-1.5}} & 77.6 & 75.3 & \textbf{\textcolor{darkred}{-2.3}} & 79.3 & 76.7 & \textbf{\textcolor{darkred}{-3.1}} \\
PCRLv2~\cite{pcrlv2}     & 67.8 & 62.8 & \textbf{\textcolor{darkred}{-5.0}} & 76.6 & 68.3 & \textbf{\textcolor{darkred}{-8.3}} & 78.2 & 73.2 & \textbf{\textcolor{darkred}{-5.0}} & 79.9 & 74.2 & \textbf{\textcolor{darkred}{-5.7}} & 81.1 & 75.1 & \textbf{\textcolor{darkred}{-6.0}} & 81.5 & 75.3 & \textbf{\textcolor{darkred}{-6.2}} \\
DINOV2~\cite{dino}       & 63.8 & 59.8 & \textbf{\textcolor{darkred}{-4.0}} & 70.6 & 67.4 & \textbf{\textcolor{darkred}{-3.2}} & 73.2 & 70.6 & \textbf{\textcolor{darkred}{-2.6}} & 77.5 & 72.0 & \textbf{\textcolor{darkred}{-5.5}} & 79.3 & 73.2 & \textbf{\textcolor{darkred}{-6.1}} & 79.8 & 74.3 & \textbf{\textcolor{darkred}{-5.5}} \\

MLVICX~\cite{mlvicx}     & 69.1 & 68.0 & \textbf{\textcolor{darkred}{-1.1}} & 74.0 & 72.6 & \textbf{\textcolor{darkred}{-1.4}} & 77.9 & 75.4 & \textbf{\textcolor{darkred}{-2.5}} & 80.5 & 77.4 & \textbf{\textcolor{darkred}{-3.1}} & \textbf{\textcolor{darkblue}{81.9}} & 78.3 & \textbf{\textcolor{darkred}{-3.6}} & \textbf{\textcolor{darkblue}{83.4}} & 78.6 & \textbf{\textcolor{darkred}{-4.8}} \\

ADAMV2~\cite{adamv2}     & 68.1 & 63.1 & \textbf{\textcolor{darkred}{-5.0}} & 74.4 & 68.4 & \textbf{\textcolor{darkred}{-6.0}} & 78.0 & 73.6 & \textbf{\textcolor{darkred}{-4.4}} & 79.7 & 74.5 & \textbf{\textcolor{darkred}{-5.2}} & 81.2 & 75.7 & \textbf{\textcolor{darkred}{-5.5}} & 81.7 & 76.3 & \textbf{\textcolor{darkred}{-5.4}} \\
OTCXR~\cite{otcxr}       & 69.6 & 65.6 & \textbf{\textcolor{darkred}{-4.0}} & 75.7 & 70.8 & \textbf{\textcolor{darkred}{-4.9}} & 78.2 & 74.4 & \textbf{\textcolor{darkred}{-3.8}} & \textbf{\textcolor{darkblue}{80.8}} & 76.3 & \textbf{\textcolor{darkred}{-4.5}} & \textbf{\textcolor{darkpurple}{81.6}} & 77.4 & \textbf{\textcolor{darkred}{-4.2}} & \textbf{\textcolor{darkpurple}{82.3}} & 77.8 & \textbf{\textcolor{darkred}{-4.5}} \\
CoBoom~\cite{coboom}     & \textbf{\textcolor{darkpurple}{71.4}} & \textbf{\textcolor{darkpurple}{70.2}} & \textbf{\textcolor{darkred}{-1.2}} & \textbf{\textcolor{darkpurple}{76.7}} & \textbf{\textcolor{darkpurple}{74.9}} & \textbf{\textcolor{darkred}{-1.8}} & \textbf{\textcolor{darkpurple}{78.4}} & \textbf{\textcolor{darkpurple}{77.0}} & \textbf{\textcolor{darkred}{-1.4}} & 79.7 & \textbf{\textcolor{darkpurple}{78.1}} & \textbf{\textcolor{darkred}{-1.6}} & 80.0 & \textbf{\textcolor{darkpurple}{78.5}} & \textbf{\textcolor{darkred}{-1.5}} & 81.1 & \textbf{\textcolor{darkpurple}{78.8}} & \textbf{\textcolor{darkred}{-2.3}} \\
\textbf{DiSSECT (Ours)}           & \textbf{\textcolor{darkblue}{73.4}} & \textbf{\textcolor{darkblue}{72.0}} & \textbf{\textcolor{darkred}{-1.4}} & \textbf{\textcolor{darkblue}{77.6}} & \textbf{\textcolor{darkblue}{76.6}} & \textbf{\textcolor{darkred}{-1.0}} & \textbf{\textcolor{darkblue}{79.1}} & \textbf{\textcolor{darkblue}{78.4}} & \textbf{\textcolor{darkred}{-0.7}} & \textbf{\textcolor{darkpurple}{80.6}} & \textbf{\textcolor{darkblue}{79.4}} & \textbf{\textcolor{darkred}{-1.2}} & 81.4 & \textbf{\textcolor{darkblue}{79.8}} & \textbf{\textcolor{darkred}{-1.6}} & 81.8 & \textbf{\textcolor{darkblue}{80.2}} & \textbf{\textcolor{darkred}{-1.6}} \\
\bottomrule
\end{tabular}
\end{table*}

\section{Experimental Setup and Evaluation}
We pre-train DiSSECT on two large-scale chest X-ray datasets—NIH ChestX-ray14~\cite{wang2017chestx} (112k radiographs) and CheXpert~\cite{irvin2019chexpert} (224k radiographs)—to learn transferable visual representations in a fully self-supervised manner. We then evaluate across classification and segmentation benchmarks.

\textbf{Pre-training Setup: }
For NIH, we follow both the official split (86k/25k train/test) and the commonly used SSL split (77k train, 11k val/test)~\cite{pcrlv1,pcrlv2,dira,adamv2}. CheXpert pretraining uses the official training set. Disease labels are never used in pre-training.

\textbf{Downstream Tasks:}
For classification, models predict chest pathologies on NIH. We assess (i) in-domain transfer (pretrain and evaluate on NIH) and (ii) cross-domain transfer (pretrain on CheXpert, evaluate on NIH). For segmentation, NIH-pretrained models are tested on pneumothorax segmentation using SIIM-ACR~\cite{siimacr2019} (12k X-rays; 70/30/30 split~\cite{wang2022multi}) and pneumonia segmentation on RSNA (29k images; converted detection boxes with 16k/5.3k/5.3k split).

\textbf{Evaluation Protocols:}
We adopt standard \textit{linear probing} (LP; frozen encoder + linear head) and \textit{semi-supervised fine-tuning} (FT; encoder + head trained jointly) at label fractions of 1–40\%. Classification is reported with AUC; segmentation uses a U-Net decoder and Dice score~\cite{wang2022multi}.

\textbf{Baseline Methods:}
We compare against supervised training, ImageNet-pretrained backbones, and SSL approaches including TransVW~\cite{transvw}, BYOL~\cite{byol}, SimSiam~\cite{simsiam}, DiRA~\cite{dira}, PCRLv2~\cite{pcrlv2}, DINOv2~\cite{dino}, MLVICX~\cite{mlvicx}, ADAMv2~\cite{adamv2}, OTCXR~\cite{otcxr}, and CoBooM~\cite{coboom}. 

\textbf{Implementation Details:}
We use PyTorch with a ResNet-18 encoder, followed by two-layer MLP projection/prediction heads (BN + ReLU). The encoder output (512-d) is projected to 2048-d then 128-d embeddings. Discrete supervision employs three VQ codebooks (coarse/medium/fine), each with 2048 entries of size 128, updated via EMA (decay 0.99) with a commitment loss ($\beta=0.25$). Models are trained for 300 epochs using LARS-SGD (lr=0.3, momentum=0.99, weight decay=$1.5 \times 10^{-6}$), with 10-epoch warmup + cosine annealing. The momentum encoder follows a cosine decay from 0.996 to 1.0. Inputs are resized to $224 \times 224$ with random crop, flip, Gaussian blur, and normalization\footnote{Code and pretrained models will be released upon paper acceptance.}.

%%%%%%%%%%%%%%%%%%%%%%%%%%%%%%%% GRAD CAM PLOTS %%%%%%%%%%%%%%%%%%%%%%%%%%%%%%

\section{Results and Analysis}
\label{sec:results}

\subsection{Low-Label Performance and Fine-Tuning Independence}

\paragraph{\textbf{Quantitative In-Domain Analysis on NIH}}  
Tables~\ref{tab:nih_orig} and~\ref{tab:nih_pcrl} compare DiSSECT with competing methods under varying label fractions. The advantages of DiSSECT are most evident in low-label regimes: on the official NIH split (Table~\ref{tab:nih_orig}), it achieves positive or near-zero LP--FT gaps at 1--5\%, indicating that the frozen encoder alone provides highly transferable features. In contrast, most baselines suffer LP drops of 3--7 AUC, even at higher supervision, showing heavy reliance on fine-tuning.  On the baseline-aligned NIH split (Table~\ref{tab:nih_pcrl}), DiSSECT again consistently outperforms prior SSL methods. At 1--5\% labels, it yields the best LP AUC and the smallest LP--FT gaps ($\leq -1.0$), while strong baselines such as PCRLv2 and ADAMv2 degrade by up to --8.3 AUC. This robustness demonstrates that DiSSECT learns embeddings less sensitive to label scarcity and downstream optimization.  
Overall, these results confirm that multi-scale discrete supervision encourages stable and semantically rich representations. By filtering shortcut correlations and view-specific noise, DiSSECT reduces the need for extensive fine-tuning and enables reliable transfer across low-label settings.

\begin{figure*}[htbp]
\centering
\begin{tabular}{cccccc}

\toprule

\multicolumn{3}{c}{\includegraphics[width=0.95\linewidth]{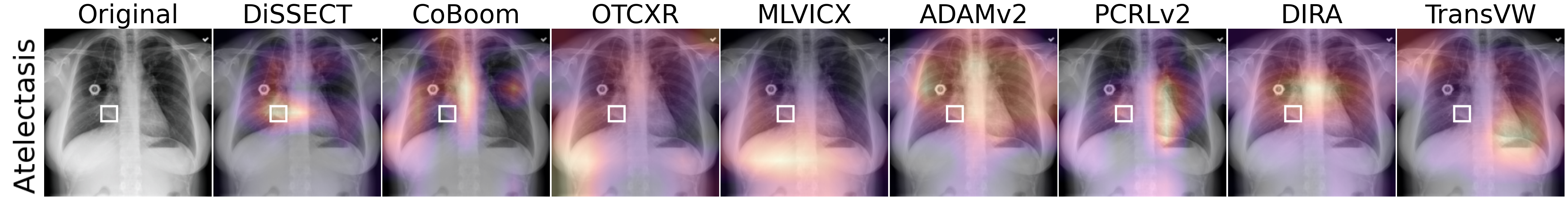}} \\
\multicolumn{3}{c}
{\includegraphics[width=0.95\linewidth]{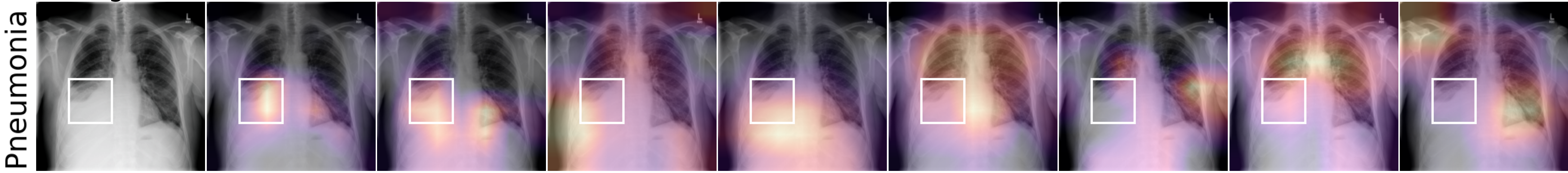}} \\

{\includegraphics[width=0.95\linewidth]{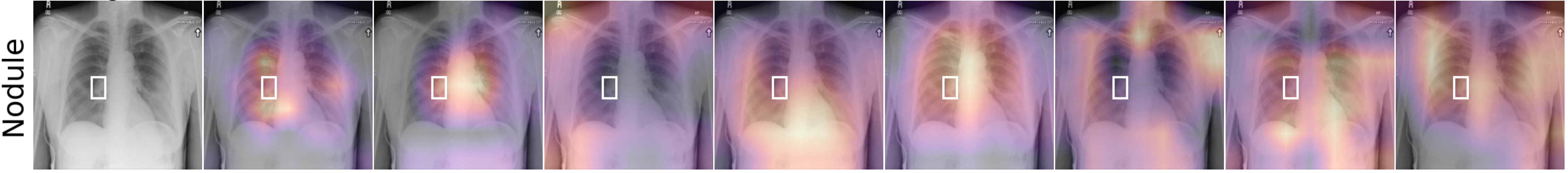}} \\

{\includegraphics[width=0.95\linewidth]{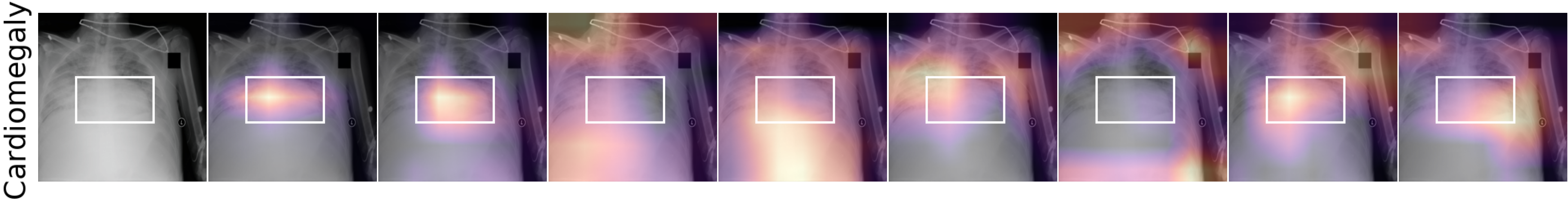}} \\

% {\includegraphics[width=0.95\linewidth]{LaTeX/plots/maps/ef3.png}} \\

{\includegraphics[width=0.95\linewidth]{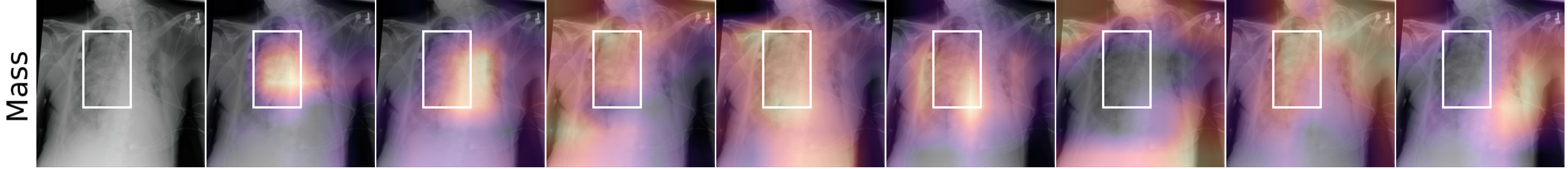}} \\

{\includegraphics[width=0.95\linewidth]{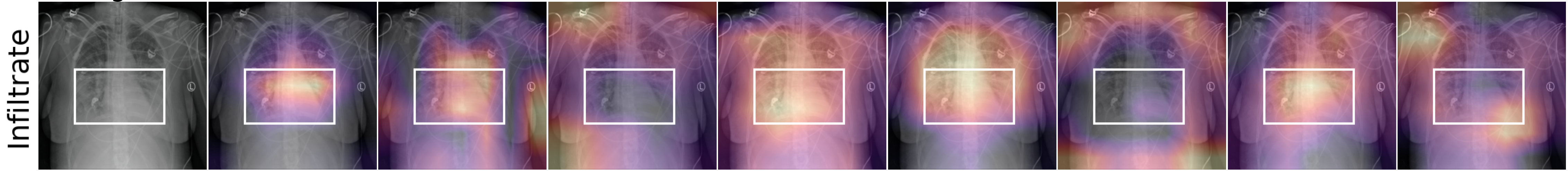}} \\

%%%%%%%%%%%%%%%%%%%%%%%%%%%%%%%

% {\includegraphics[width=0.95\linewidth]
% {LaTeX/plots/maps/d_12.png}} \\
% {\includegraphics[width=0.95\linewidth]{LaTeX/plots/maps/pn18.png}} \\

% {\includegraphics[width=0.95\linewidth]{LaTeX/plots/maps/nd1.png}} \\

% {\includegraphics[width=0.95\linewidth]{LaTeX/plots/maps/cd_25.png}} \\

% {\includegraphics[width=0.95\linewidth]{LaTeX/plots/maps/ef5.png}} \\

% {\includegraphics[width=0.95\linewidth]{LaTeX/plots/maps/m5.png}} \\

% \toprule
\bottomrule

\end{tabular}
\caption{GradCAM visualizations from pretrained encoders on NIH (baseline split) without any fine-tuning. DiSSECT consistently localizes relevant regions accurately compared to others, indicating stronger semantic alignment during pre-training.}

\label{fig_cam:gradcam}
\vspace{-14pt} 
\end{figure*}

\begin{figure*}[htbp]
    \centering
    % Top row
    \begin{subfigure}[b]{0.23\textwidth}
        \centering
        \includegraphics[width=\linewidth]{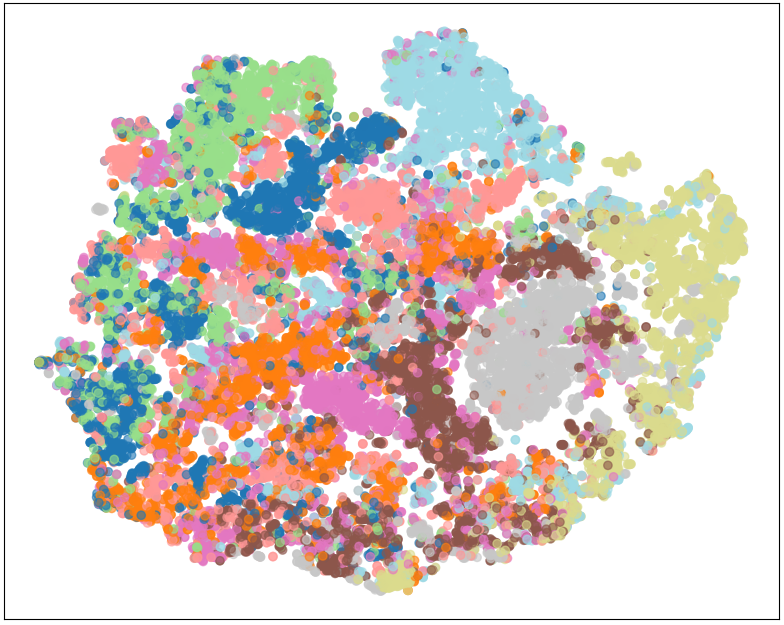}
        \caption*{\textbf{TransVW}}
    \end{subfigure}
    \begin{subfigure}[b]{0.23\textwidth}
        \centering
        \includegraphics[width=\linewidth]{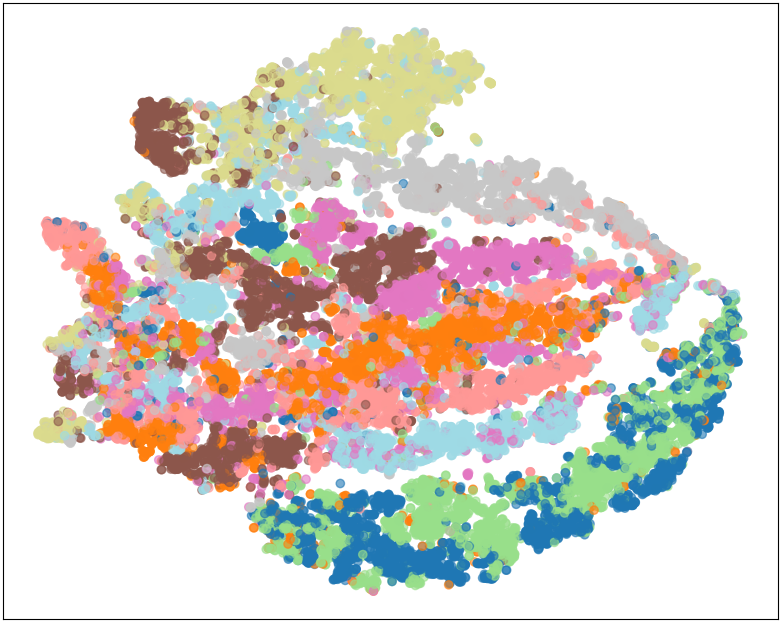}
        \caption*{\textbf{DIRA}}
    \end{subfigure}
    \begin{subfigure}[b]{0.23\textwidth}
        \centering
        \includegraphics[width=\linewidth]{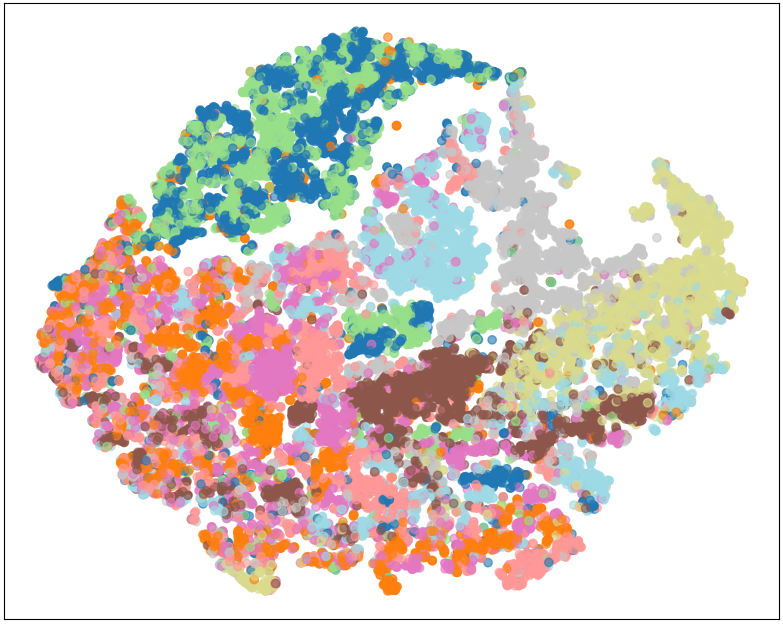}
        \caption*{\textbf{MLVICX}}
    \end{subfigure}
    \begin{subfigure}[b]{0.23\textwidth}
        \centering
        \includegraphics[width=\linewidth]{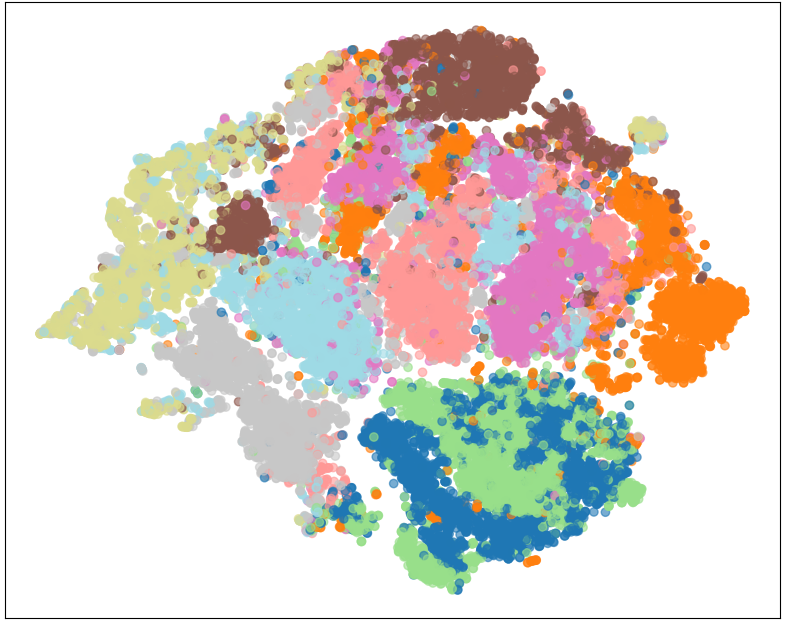}
        \caption*{\textbf{OTCXR}}
    \end{subfigure}

    \vspace{1em}

    % Bottom row
    \begin{subfigure}[b]{0.23\textwidth}
        \centering
        \includegraphics[width=\linewidth]{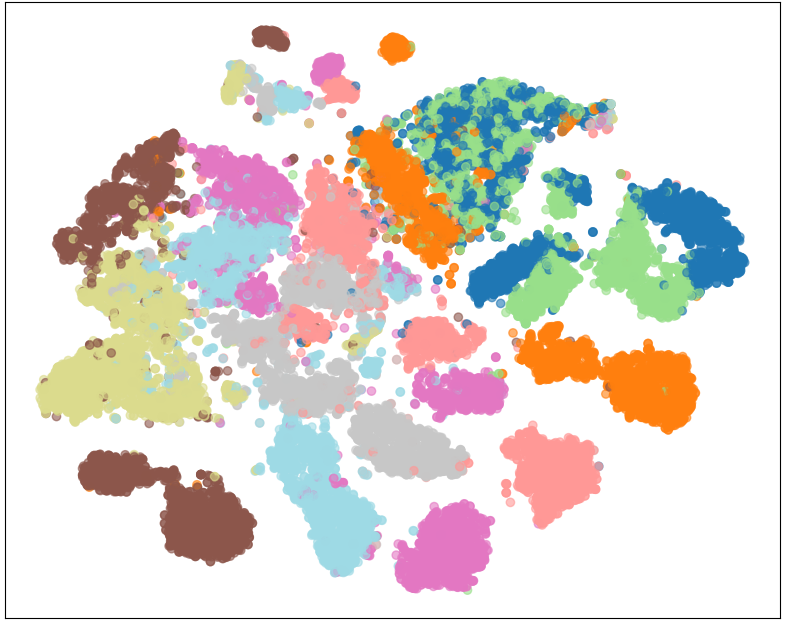}
        \caption*{\textbf{PCRLv2}}
    \end{subfigure}
    \begin{subfigure}[b]{0.23\textwidth}
        \centering
        \includegraphics[width=\linewidth]{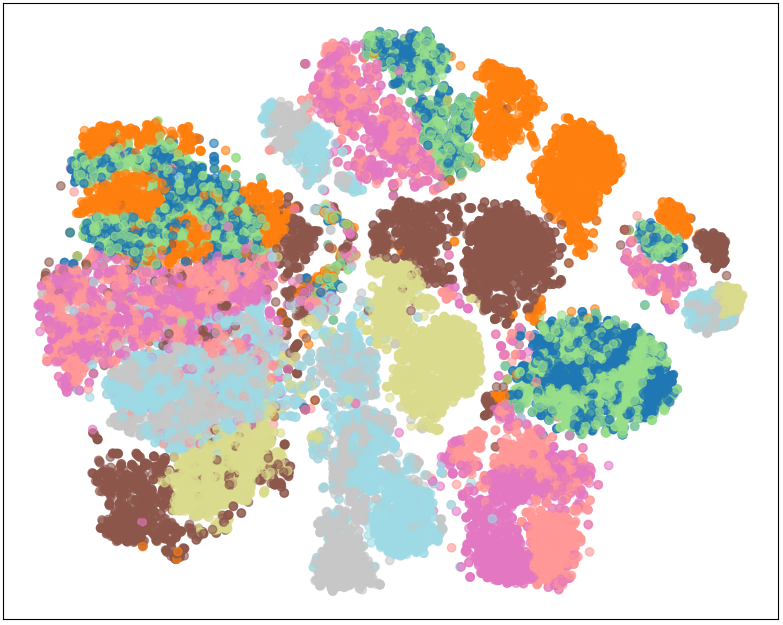}
        \caption*{\textbf{CoBoom}}
    \end{subfigure}
    \begin{subfigure}[b]{0.23\textwidth}
        \centering
        \includegraphics[width=\linewidth]{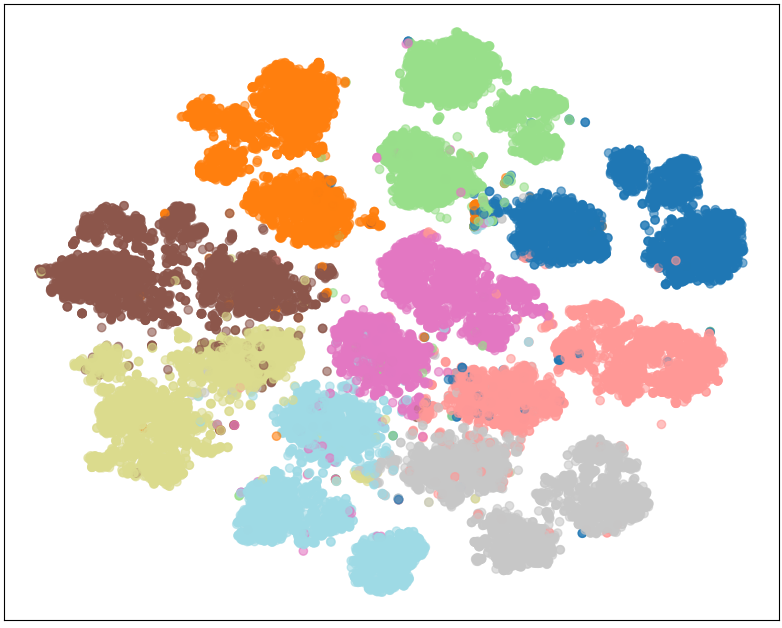}
        \caption*{\textbf{ADAMv2}}
    \end{subfigure}
    \begin{subfigure}[b]{0.23\textwidth}
        \centering
        \includegraphics[width=\linewidth]{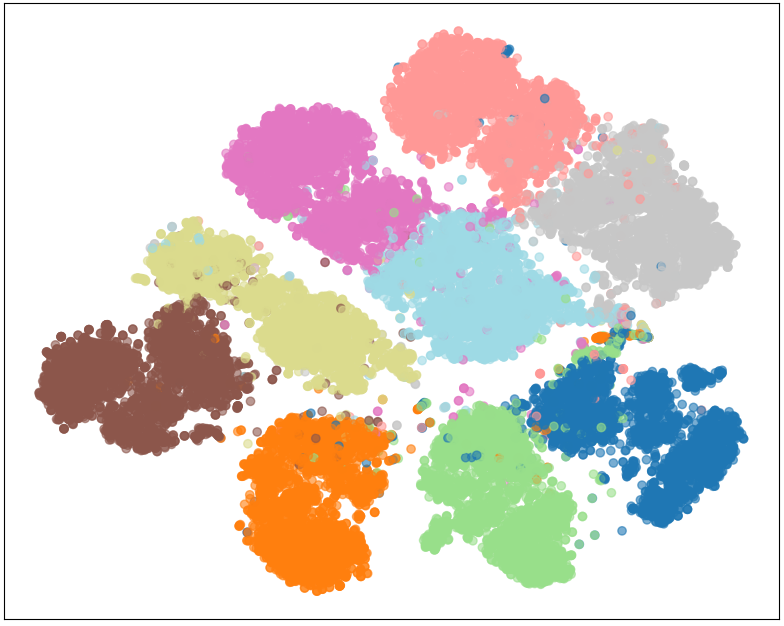}
        \caption*{\textbf{DiSSECT}}
    \end{subfigure}

    \caption{T-SNE visualization of region-level embeddings from encoders pretrained with different SSL methods. Each point represents a $3 \times 3$ grid patch from an input image, and colors denote patch positions (e.g., top-left, center, bottom-right) consistently across all models. Clear separation by color indicates that the encoder captures spatially coherent and anatomically consistent representations.}
    \label{fig:tsne}
\end{figure*}

\begin{table*}[htbp]
\centering
\caption{Performance of SSL methods pretrained on Chexpert and evaluated on NIH (baseline split) under transfer learning.
Best and second-best FT/LP scores are highlighted in \textcolor{darkblue}{\textbf{blue}} and \textcolor{darkpurple}{\textbf{purple}}, respectively. LP - FT ($\Delta$) differences are in \textcolor{darkred}{\textbf{red}}.}
\label{tab:official_ft_lp_diff}
\setlength{\tabcolsep}{4pt}
\begin{tabular}{l|ccc|ccc|ccc|ccc|ccc|ccc}
\toprule
\multirow{2}{*}{\textbf{Method}} 
& \multicolumn{3}{c|}{\textbf{1\%}} 
& \multicolumn{3}{c|}{\textbf{5\%}} 
& \multicolumn{3}{c|}{\textbf{10\%}} 
& \multicolumn{3}{c|}{\textbf{20\%}} 
& \multicolumn{3}{c|}{\textbf{30\%}} 
& \multicolumn{3}{c}{\textbf{40\%}} \\
 & FT & LP & $\Delta$ & FT & LP & $\Delta$ & FT & LP & $\Delta$ & FT & LP & $\Delta$ & FT & LP & $\Delta$ & FT & LP & $\Delta$ \\
\midrule
Random Init.     & 58.9 & 56.6 & \textbf{\textcolor{darkred}{-2.3}} & 65.8 & 60.3 & \textbf{\textcolor{darkred}{-5.5}} & 68.1 & 63.1 & \textbf{\textcolor{darkred}{-5.0}} & 71.5 & 64.2 & \textbf{\textcolor{darkred}{-7.3}} & 73.4 & 64.6 & \textbf{\textcolor{darkred}{-8.8}} & 75.4 & 65.0 & \textbf{\textcolor{darkred}{-10.4}} \\
ImageNet Init.   & 63.9 & 62.2 & \textbf{\textcolor{darkred}{-5.0}} & 70.4 & 65.4 & \textbf{\textcolor{darkred}{-5.3}} & 73.5 & 68.2 & \textbf{\textcolor{darkred}{-5.6}} & 76.2 & 70.6 & \textbf{\textcolor{darkred}{-7.4}} & 78.5 & 71.1 & \textbf{\textcolor{darkred}{-5.0}} & 79.0 & 71.9 & \textbf{\textcolor{darkred}{-5.4}} \\
TransVW~\cite{transvw}   & 62.8  & 60.6 & \textbf{\textcolor{darkred}{-2.2}} & 66.9 & 63.8 & \textbf{\textcolor{darkred}{-3.1}} & 69.7 & 67.2 & \textbf{\textcolor{darkred}{-2.5}} & 73.8 & 71.0 & \textbf{\textcolor{darkred}{-2.8}} & 75.7 & 72.4 & \textbf{\textcolor{darkred}{-3.3}} & 77.3 & 74.4 & \textbf{\textcolor{darkred}{-2.9}} \\
BYOL~\cite{byol}        & 66.7 & 65.4 & \textbf{\textcolor{darkred}{-1.3}} & 72.5 & 70.1 & \textbf{\textcolor{darkred}{-2.4}} & 74.5 & 72.6& \textbf{\textcolor{darkred}{-1.9}} & 77.2 & 74.7 & \textbf{\textcolor{darkred}{-2.5}} & 79.3 & 76.1& \textbf{\textcolor{darkred}{-3.2}} & 79.8 & 76.5 & \textbf{\textcolor{darkred}{-3.3}} \\
SimSiam~\cite{simsiam}   & 64.4 & 62.3 & \textbf{\textcolor{darkred}{-2.1}} & 69.2 & 66.8 & \textbf{\textcolor{darkred}{-2.4}} & 72.5 & 70.3  & \textbf{\textcolor{darkred}{-2.2}} & 76.6 & 73.5 & \textbf{\textcolor{darkred}{-3.1}} & 78.3 & 74.6 & \textbf{\textcolor{darkred}{-3.7}} & 78.9 & 75.3 & \textbf{\textcolor{darkred}{-3.6}} \\
DIRA~\cite{dira}         & OT & -- & \textbf{\textcolor{darkred}{--}} & -- & -- & \textbf{\textcolor{darkred}{--}} & -- & -- & \textbf{\textcolor{darkred}{--}} & -- & -- & \textbf{\textcolor{darkred}{--}} & -- & -- & \textbf{\textcolor{darkred}{--}} & -- & -- & \textbf{\textcolor{darkred}{--}} \\
PCRLv2~\cite{pcrlv2}     & 66.8  & 64.7 & \textbf{\textcolor{darkred}{-2.1}} & 71.9 &69.8 & \textbf{\textcolor{darkred}{-2.1}} & 77.2 & 74.4 & \textbf{\textcolor{darkred}{-2.8}} & 79.4 & 75.1 & \textbf{\textcolor{darkred}{-4.3}} & 80.5 & 75.9 & \textbf{\textcolor{darkred}{-4.6}} & \textbf{\textcolor{darkpurple}{81.5}} & 76.1 & \textbf{\textcolor{darkred}{-5.4}} \\

MLVICX~\cite{mlvicx}& 69.4 & 67.7 & \textbf{\textcolor{darkred}{-1.7}} & 74.7 & 72.8 & \textbf{\textcolor{darkred}{-1.9}} & 77.5 & 75.2 & \textbf{\textcolor{darkred}{-2.3}} & 78.9 & 76.7 & \textbf{\textcolor{darkred}{-2.2}} & 79.9 &76.9 & \textbf{\textcolor{darkred}{-3.0}} & 80.3 & 77.3 & \textbf{\textcolor{darkred}{-3.0}} \\

ADAMV2~\cite{adamv2}     & 66.6  & 62.1 & \textbf{\textcolor{darkred}{-4.5}} & 71.0 & 68.2 & \textbf{\textcolor{darkred}{-2.8}} & 75.8 & 70.2 & \textbf{\textcolor{darkred}{-5.6}} & 78.8 & 72.8 & \textbf{\textcolor{darkred}{-6.0}} & 80.0 & 73.5 & \textbf{\textcolor{darkred}{-6.5}} & 80.3 & 74.8 & \textbf{\textcolor{darkred}{-5.5}} \\
OTCXR~\cite{otcxr}       & 67.4 & 63.9 & \textbf{\textcolor{darkred}{-3.5}} & 73.5 & 69.3 & \textbf{\textcolor{darkred}{-4.2}} & \textbf{\textcolor{darkblue}{79.4}} & 73.2 & \textbf{\textcolor{darkred}{-6.2}} & 78.3 & 74.6 & \textbf{\textcolor{darkred}{-3.7}} & 79.6 & 76.3 & \textbf{\textcolor{darkred}{-3.3}} & 80.5 & 77.4 & \textbf{\textcolor{darkred}{-3.1}} \\
CoBoom~\cite{coboom}     & \textbf{\textcolor{darkpurple}{69.7}} & \textbf{\textcolor{darkpurple}{68.3}} & \textbf{\textcolor{darkred}{-1.4}} & \textbf{\textcolor{darkpurple}{75.2}} & \textbf{\textcolor{darkpurple}{74.0}} & \textbf{\textcolor{darkred}{-1.2}} & 77.4 & \textbf{\textcolor{darkpurple}{76.2}} & \textbf{\textcolor{darkred}{-1.2}} & \textbf{\textcolor{darkpurple}{79.4}} & \textbf{\textcolor{darkpurple}{77.0}} & \textbf{\textcolor{darkred}{-2.4}} & \textbf{\textcolor{darkblue}{81.2}} & \textbf{\textcolor{darkpurple}{77.6}} & \textbf{\textcolor{darkred}{-3.6}} & 81.0 & \textbf{\textcolor{darkpurple}{78.0}} & \textbf{\textcolor{darkred}{-3.0}} \\
\textbf{DiSSECT (Ours) }          & \textbf{\textcolor{darkblue}{71.4}} & \textbf{\textcolor{darkblue}{71.0}} & \textbf{\textcolor{darkred}{-0.4}} & \textbf{\textcolor{darkblue}{76.4}} & \textbf{\textcolor{darkblue}{76.1}} & \textbf{\textcolor{darkred}{-0.3}} & \textbf{\textcolor{darkpurple}{78.2}} & \textbf{\textcolor{darkblue}{78.0}} & \textbf{\textcolor{darkred}{-0.2}} & \textbf{\textcolor{darkblue}{79.6}} & \textbf{\textcolor{darkblue}{79.0}} & \textbf{\textcolor{darkred}{-0.6}} & \textbf{\textcolor{darkpurple}{80.8}} & \textbf{\textcolor{darkblue}{79.6}} & \textbf{\textcolor{darkred}{-1.2}} & \textbf{\textcolor{darkblue}{81.6}} & \textbf{\textcolor{darkpurple}{80.0}} & \textbf{\textcolor{darkred}{-1.6}} \\
\bottomrule
\end{tabular}
\end{table*}

%%%%%%%%%%%%%%%%%%%%% Segmentation %%%%%%%%%%%%%%%%%%%%%%%%

\begin{table*}[t]
\centering
\caption{Segmentation Dice scores [\%] on SIIM and RSNA across label fractions under LP and FT.}
\setlength{\tabcolsep}{5pt}
\begin{tabular}{l|cc|cc|cc|cc||cc|cc|cc|cc}
\toprule
\multirow{3}{*}{\textbf{Method}} & \multicolumn{8}{c||}{SIIM} & \multicolumn{8}{c}{RSNA} \\
\cmidrule(lr){2-9} \cmidrule(lr){10-17}
 & \multicolumn{2}{c|}{1\%} & \multicolumn{2}{c|}{5\%} & \multicolumn{2}{c|}{10\%} & \multicolumn{2}{c||}{100\%} 
 & \multicolumn{2}{c|}{1\%} & \multicolumn{2}{c|}{5\%} & \multicolumn{2}{c|}{10\%} & \multicolumn{2}{c}{100\%} \\
 & FT & LP & FT & LP & FT & LP & FT & LP & FT & LP & FT & LP & FT & LP & FT & LP \\
\midrule
Random Init.     & 16.3 & 11.7 & 21.6 & 16.7 & 28.6 & 19.8 & 54.3 & 48.3 & 18.1 & 9.8 & 42.9 & 30.8 & 47.2 & 39.5 & 56.9 & 51.6 \\
ImageNet Init.   & 26.1 & 23.3 & 49.5 & 46.8 & 51.6 & 48.3 & 66.5 & 63.7 & 37.2 & 28.9 & 51.6 & 47.0 & 53.7 & 51.6 & 72.4 & 68.9 \\
% MGCA~\cite{wang2022multi}          & - & - & - & - & 59.3 & - & 64.2 & - & - & - & - & - & 68.3 & - & 69.8 & - \\
TransVW~\cite{transvw}                   & 63.3 & 60.1 & 63.7 & 61.7 & 65.6 & 63.2 & 68.7 & 65.2 & 72.8 & 71.4 & 74.1 & 70.7 & 74.9 & 73.5 & 76.1 & 73.8 \\
% DINOv2~\cite{dino}                 & - & - & - & - & 58.1 & - & - & - & - & - & - & - & - & - & - & - \\
DIRA~\cite{dira}                   & 62.7 & 60.3 & 63.2 & 61.5 & 68.6 & 66.9 & 70.6 & 68.9 & 68.7 & 66.4 & 71.0 & 69.4 & 72.3 & 70.1 & 73.1 & 70.9 \\
PCRLv2~\cite{pcrlv2}                   & 70.9 & 69.7 & 72.3 & 71.9 & 72.8 & 72.0 & 73.8 & 72.4 & 72.2 & 71.5 & 73.9 & 72.5 & 75.5 & 74.7 & 76.9 & 75.7 \\
MLVICX~\cite{mlvicx}                   & 67.1 & 69.1 & 70.1 & 71.2 & 71.3 & 72.1 & 71.6 & 72.7 & 72.6 & 71.0 & 73.2 & 72.2 & 74.6 & 74.0 & 77.3 & 75.9 \\
ADAMv2~\cite{adamv2}                   & 56.3 & 51.4 & 69.7 & 67.8 & 72.1 & 70.3 & 72.9 & 71.4 & \textbf{\textcolor{darkpurple}{73.6}} & 72.0 & \textbf{\textcolor{darkpurple}{74.6}} & 73.0 & 75.2 & 75.1 & 75.8 & 75.1 \\
OTCXR~\cite{otcxr}                   & 68.9 & 70.1 & 67.3 & 69.5 & 70.6 & 71.3 & 72.6 & 71.8 & 71.4 & 71.1 & 73.3 & 72.4 & \textbf{\textcolor{darkpurple}{75.6}} & 73.1 & 77.9 & 74.5 \\
CoBoom~\cite{coboom}              & \textbf{\textcolor{darkpurple}{71.7}} & \textbf{\textcolor{darkpurple}{71.2}} & \textbf{\textcolor{darkpurple}{72.2}} & \textbf{\textcolor{darkpurple}{72.4}} & \textbf{\textcolor{darkpurple}{72.6}} & \textbf{\textcolor{darkpurple}{72.9}} & \textbf{\textcolor{darkpurple}{74.3}} & \textbf{\textcolor{darkpurple}{73.6}} & 72.0 & \textbf{\textcolor{darkpurple}{72.3}} & 74.1 & \textbf{\textcolor{darkpurple}{74.4}} & 75.1 & \textbf{\textcolor{darkpurple}{75.3}} & \textbf{\textcolor{darkblue}{78.3}} & \textbf{\textcolor{darkpurple}{76.7}} \\
\textbf{DiSSECT (Ours)}           & \textbf{\textcolor{darkblue}{71.9}} & \textbf{\textcolor{darkblue}{72.2}} & \textbf{\textcolor{darkblue}{73.4}} & \textbf{\textcolor{darkblue}{73.1}} & \textbf{\textcolor{darkblue}{73.7}} & \textbf{\textcolor{darkblue}{73.4}} & \textbf{\textcolor{darkblue}{74.9}} & \textbf{\textcolor{darkblue}{73.9}} & \textbf{\textcolor{darkblue}{74.4}} & \textbf{\textcolor{darkblue}{74.8}} & \textbf{\textcolor{darkblue}{75.1}} & \textbf{\textcolor{darkblue}{75.9}} & \textbf{\textcolor{darkblue}{77.3}} & \textbf{\textcolor{darkblue}{76.6}} & \textbf{\textcolor{darkpurple}{78.1}} & \textbf{\textcolor{darkblue}{77.9}} \\
\bottomrule
\end{tabular}
\label{tab:segmentation_results}
\end{table*}

\paragraph{\textbf{Qualitative Analysis on NIH}}  
Figure~\ref{fig_cam:gradcam} shows GradCAM maps from frozen encoders on NIH chest X-rays. DiSSECT consistently produces compact activations centered within the annotated pathology regions, accurately localizing abnormalities such as the cardiac silhouette in Cardiomegaly, pleural consolidations in Pneumonia, and small localized lesions in Nodule and Mass. In contrast, most baselines exhibit diffuse or misplaced responses: CoBoom and ADAMv2 partially capture relevant areas but spread attention widely, while OTCXR, MLVICX, PCRLv2, DIRA, and TransVW often highlight non-diagnostic regions (e.g., chest wall, diffuse lung fields).  
This difference is especially clear in subtle cases such as Nodules, where DiSSECT is the only method that tightly localizes the lesion within the annotated region, while others activate broad lung textures. These visualizations confirm that multi-scale discrete supervision enables DiSSECT to focus on clinically meaningful structures even without fine-tuning. By contrast, baselines show weaker alignment between pre-trained features and pathological regions, underscoring DiSSECT’s advantage in learning semantically grounded and transferable representations.

\paragraph{\textbf{Anatomical Structure Embedding Visualization}} 
To demonstrate DiSSECT’s ability to generalize without relying on anatomy-specific supervision, we analyze region-level embeddings via T-SNE, revealing how multi-scale quantization naturally organizes features into semantically meaningful clusters. Each chest X-ray is divided into a $3 \times 3$ grid, and patch-level features from frozen encoders are projected without fine-tuning. DiSSECT forms distinct clusters that align with anatomical regions, despite no supervision. This structure arises from vector quantization: recurring local patterns are mapped to stable codebook vectors, providing discrete anchors.  
In contrast, CoBoom and PCRLv2 show overlapping clusters, indicating weaker anatomical separation. ADAMv2~\cite{adamv2}, which introduces explicit anatomical supervision through part–whole hierarchy objectives (localizability, composability, decomposability), produces tightly compact clusters. While these clusters reflect strong prior-driven organization, their rigidity may reduce adaptability across datasets with anatomical variability. By comparison, DiSSECT achieves anatomical consistency as an emergent property of quantization, requiring no annotations or handcrafted supervision. This emergent alignment not only validates the effectiveness of discrete supervision but also explains why DiSSECT maintains strong transferability and requires minimal fine-tuning in downstream tasks.

\subsection{Cross-Domain Transfer: Cross-Dataset Classification and Segmentation}

\paragraph{\textbf{Cross-Dataset Classification (CheXpert $\rightarrow$ NIH)}} 
Table~\ref{tab:official_ft_lp_diff} shows that DiSSECT consistently achieves the best or near-best performance across all label fractions under both FT and LP. Its LP–FT gap remains exceptionally small (–0.2 to –1.6 AUC), while prior SSL baselines such as PCRLv2, ADAMv2, and OTCXR suffer much larger drops (–4 to –6 AUC). This indicates that DiSSECT produces stable, transferable representations that generalize across datasets without requiring heavy fine-tuning. The advantage is most pronounced in the low-label regime, where DiSSECT outperforms all baselines at 1\% and 5\% supervision, confirming that its discrete bottleneck captures robust semantics rather than dataset-specific correlations.

\paragraph{\textbf{Segmentation Transfer (NIH $\rightarrow$ SIIM and RSNA)}} 
Table~\ref{tab:segmentation_results} reports segmentation Dice scores under LP and FT. DiSSECT achieves the strongest or comparable results in nearly all settings, particularly under 1\% and 5\% supervision, where it surpasses both anatomy-aware (ADAMv2) and VQ-based (CoBoom) baselines. On SIIM pneumothorax, DiSSECT achieves 72.2–73.4 Dice with only 1–5\% labels, while CoBoom lags behind. On RSNA lung opacity, DiSSECT again leads, with 74.8/75.9 Dice under 1–5\% labels, outperforming all baselines. Even at full supervision, DiSSECT remains competitive, showing that its advantage extends beyond the low-label setting. These results highlight that multi-scale discrete supervision enables embeddings that preserve spatial semantics and generalize across domains and tasks, offering efficient adaptation to new clinical settings without reliance on extensive fine-tuning.

\begin{table}[t]
\centering
\caption{Ablation study results on NIH (AUC) and SIIM (Dice) using 1\% and 5\% labeled data.}

\label{tab:ablation_lp_ft}
\setlength{\tabcolsep}{2pt}
\begin{tabular}{l|cc|cc||cc|cc}
\toprule
\multirow{3}{*}{\textbf{Ablation Variant}} 
& \multicolumn{4}{c||}{\textbf{NIH}} & \multicolumn{4}{c}{\textbf{SIIM}} \\
\cmidrule(lr){2-5} \cmidrule(lr){6-9}

& \multicolumn{2}{c|}{1\%} & \multicolumn{2}{c||}{5\%} 
& \multicolumn{2}{c|}{1\%} & \multicolumn{2}{c}{5\%} \\

& FT & LP & FT & LP & FT &  LP & FT & LP \\
\midrule
Full DiSSECT                           &  73.4    &  72.0    &  77.6    &  76.6    &  71.9    &  72.2 & 73.4 & 73.1     \\
w/o SERF (concat fusion)              &  69.0    &  53.0    &   74.2   &  57.8    & 68.0  &  57.9  & 70.0  &  59.3   \\
w/o Proj. Head (after SERF)      &  72.7   &  71.3    &  76.9    & 75.8     & 71.0     &  71.7  & 73.0 & 72.3   \\
w Coarse Quantization Only                &  71.7     &  69.8    & 75.4     & 73.7     &  70.4     &  69.8 & 71.7 & 72.0    \\
w Median Quantization only               &   70.0   & 67.7     & 73.3     &  70.6    &  70.9    & 71.3   & 72.1 & 71.5 \\
w Fine Quantization only                 &   71.1   & 69.6     &  75.3    &  74.6    &  68.8    &  67.2 & 70.6 & 69.1    \\
Only $h_\phi$ ($\mathcal{L}_{reg}(h_\theta , h_\phi)$)   & 68.0     & 65.1       &  73.3    & 71.7     & 63.4     &  60.1 & 68.7 & 64.3   \\
Only $q_t$ ($\mathcal{L}_{reg}(h_\theta , q_t)$)      &  70.1    &  71.2    &  74.8    &  74.9    &  69.3   & 68.1 &  72.1 & 71.4   \\
No Momentum Update                     &  21.3    & 19.2  &  27.4   &  23.7    &  13.7   &  10.4  & 19.6 & 15.7     \\
\bottomrule
\end{tabular}
\end{table}

\section{Ablation Studies}
We perform ablations on NIH (classification) and SIIM (segmentation) at 1\% and 5\% label fractions to evaluate the role of key components in DiSSECT (Table~\ref{tab:ablation_lp_ft}).

\paragraph{\textbf{Effect of SERF Fusion}}
Row 2 of Table~\ref{tab:ablation_lp_ft} replaces SERF with a naive concatenation of $q_\phi$ and $h_\phi$, which is then passed through the projection head. This causes severe degradation—NIH (1\% labels) LP AUC drops from 72.0 to 53.0, and SIIM (5\%) Dice from 72.2 to 57.9—showing that SERF’s structured fusion is essential for anchoring supervision on compressible, anatomically meaningful regions while suppressing irrelevant variations.

\paragraph{\textbf{Effect of the Projection Head}}
Row 3 of Table~\ref{tab:ablation_lp_ft} removes the projection head, aligning $h_\theta$ directly with $k_\phi$. This yields modest but consistent drops (e.g., NIH 5\% LP AUC 76.6$\longrightarrow$75.8, SIIM 5\% Dice 73.1$\longrightarrow$72.3), indicating that the projection head helps decouple spaces, ensuring smoother gradients and stronger generalization.

\paragraph{\textbf{Impact of Multi-Scale Quantization}} 
We ablate DiSSECT’s coarse, medium, and fine quantizers by retaining only one scale at a time (Rows 4–6). Coarse-only shows moderate drops, while fine-only collapses (SIIM 1\% Dice 72.2$\rightarrow$67.2), and medium-only is similarly suboptimal. No single scale suffices; multi-scale fusion is essential to capture both global structure and local pathology.

\paragraph{\textbf{Contribution of Semantic and Discrete Targets}}  
Rows 7–8 test the dual alignment loss $\mathcal{L}_{\text{sim}}$ by isolating each target. Using only $h_\phi$ causes large drops (NIH 1\% LP 72.0$\rightarrow$65.1; SIIM 5\% Dice 72.2$\rightarrow$60.1). Using only $q_t$ performs better but still below the full model. Together, $h_\phi$ ensures semantic consistency while $q_t$ captures compressible structure, and their joint use provides the strongest signal for generalizable representation learning.

\paragraph{\textbf{Effect of Momentum Update Mechanism}}  
Row 9 removes the momentum encoder, forcing both branches to share weights without temporal smoothing. This leads to training collapse (e.g., NIH 1\% FT 73.4$\rightarrow$21.3; SIIM 5\% Dice 73.1$\rightarrow$15.7), as rapidly changing supervision prevents codebook convergence. Stable, slowly evolving targets are essential for quantization to form consistent clusters, and the momentum update provides this stability, making it indispensable for discrete supervision.

\section{Discussion and Limitations}
\textbf{Comparison with State-of-the-Art:}  
A recurring limitation of many SSL approaches in medical imaging is that they overlook the inherent anatomical similarities shared across radiographs, instead relying on auxiliary components or supervision. For example, PCRLv2~\cite{pcrlv2} and DiRA~\cite{dira} employ adversarial or multi-stage objectives, while ADAMv2~\cite{adamv2} encodes part–whole hierarchies through explicit anatomical priors and multi-branch designs. Although effective, these strategies increase pipeline complexity and depend on handcrafted supervision or architectural modifications. DINOv2~\cite{dino}, while powerful as a self-distilled vision transformer, remains task-agnostic and does not explicitly capture structural consistency in medical images. Similarly, methods such as MLVICX~\cite{mlvicx} and OTCXR~\cite{otcxr} regularize features using variance–covariance or transport-based objectives, yet still underutilize recurring anatomical structure across datasets.  

In contrast, DiSSECT builds upon the intuition of CoBooM~\cite{coboom}, which first demonstrated the potential of VQ-based discrete supervision in medical SSL. While CoBooM improved stability and implicit anatomical grouping through quantization, it relied on a single-scale codebook. DiSSECT generalizes this idea with multi-scale vector quantization and SERF fusion, enabling compressible and repeatable features across coarse, medium, and fine semantic levels \emph{without requiring explicit priors, auxiliary branches, or multi-stage training}.  

This design yields two major benefits: (i) features become spatially grounded and semantically stable, supporting robust transfer across datasets, and (ii) quantization imposes an inductive bias toward abnormality localization, as reflected in Grad-CAM maps (Fig.~\ref{fig_cam:gradcam}) that consistently highlight clinically relevant regions without supervision. Empirically, DiSSECT achieves the smallest LP–FT gaps among all baselines (–0.2 to –1.6 AUC in Tables~\ref{tab:nih_orig} and~\ref{tab:nih_pcrl}), compared to drops of –4 AUC or more in PCRLv2, ADAMv2, and OTCXR. On cross-domain segmentation (NIH$\rightarrow$SIIM/RSNA), DiSSECT also surpasses competing SSL methods, particularly under 1–5\% labels, with gains of 2–4 Dice points (Table~\ref{tab:segmentation_results}). Overall, by encoding anatomical similarity implicitly through a discrete bottleneck, DiSSECT achieves both efficiency and clinical relevance: lightweight training, reduced fine-tuning dependence, robust generalization across cohorts, and strong pathology alignment without explicit supervision.

\textbf{Limitations:}  (i) \emph{Codebook Capacity vs. Clinical Diversity:} A fixed-size codebook may underrepresent subtle or rare pathologies, collapsing clinically distinct cues into shared tokens; adaptive or uncertainty-aware expansion could mitigate this.  
(ii) \emph{Quantization Stability:} Removing momentum updates leads to training collapse, underscoring VQ’s sensitivity to unstable targets; more principled stabilizers (e.g., regularized assignment or Bayesian smoothing) could enhance robustness.  
(iii) \emph{Discrete Bottleneck vs. Pathology Continuity:} Discretization enforces hard partitions, which may oversimplify continuous disease progressions; hybrid discrete–continuous embeddings could better capture clinical spectra.  
(iv) \emph{Pretraining Overhead:} Multi-scale VQ adds moderate cost, though still lighter than multi-stage pipelines (e.g., DiRA, ADAM), and inference remains unaffected.

\section{Conclusion}
We introduced DiSSECT, a lightweight framework for discrete SSL in medical imaging. By combining multi-scale VQ with a SERF fusion module, it learns compressible, anatomically relevant features without requiring labels or adding major complexity. Across NIH, CheXpert, RSNA, and SIIM benchmarks, DiSSECT achieves strong in-domain and cross-domain results, particularly in low-label settings and even without fine-tuning. Both quantitative and qualitative evidence confirm that discrete supervision enhances spatial and semantic consistency. Future work may extend DiSSECT to other modalities, adaptive quantization, or multi-modal settings.

\bibliographystyle{IEEEtran}
\bibliography{refer}

\end{document}